\documentclass[runningheads]{llncs}
\usepackage{graphicx}
\usepackage{booktabs}
\usepackage{hyperref}
\usepackage{amsmath}
\usepackage{amssymb}
\usepackage{multirow}
\usepackage{pifont}

\usepackage[table, dvipsnames]{xcolor}
\usepackage[table,xcdraw]{xcolor}
\usepackage{colortbl}
\usepackage[normalem]{ulem}
\useunder{\uline}{\ul}{}

\begin{document}
\title{SCALPEL: Semantic Cross-modal Alignment via LLM-Powered Encoder Learning for Medical Vision-Language Representation}
\author{Yunzhan Fu\inst{1}\textsuperscript{$*$} \and Enyu Bao\inst{1}\textsuperscript{$*$} \and
Xiangyu Shen\inst{1} \and
Yihao Wu\inst{1} \and
Chunbo Jiang\inst{1} \and
Fangli Guan\inst{1}\textsuperscript{$\dagger$} \and
Liqi Yan\inst{1}\textsuperscript{$\dagger$}}
\authorrunning{Y. Fu et al.}
%
\institute{School of Computer Science, Hangzhou Dianzi University, China \\
\email{yunzhanfu@hdu.edu.cn}, \email{23050316@hdu.edu.cn}, \email{sonaengji@gmail.com}, \email{wuyihaowyh@gmail.com}, \email{24050828@hdu.edu.cn}, \email{fangli.guan@hdu.edu.cn}, \email{lqyan18@fdu.edu.cn}\\ \textsuperscript{$*$} Equal contribution. \textsuperscript{$\dagger$} Corresponding author.}
\maketitle  
\begin{abstract}
Vision-language pre-training (VLP) serves as a cornerstone for medical multimodal representation learning. However, existing medical VLP frameworks are often constrained by the limited context windows and shallow representational capacities of lightweight text encoders when processing lengthy, terminology-dense clinical reports. While integrating medical large language models (LLMs) offers unprecedented clinical reasoning capabilities, it introduces three major bottlenecks: (i) the anisotropic representational collapse of generative LLMs under standard contrastive objectives, (ii) the prohibitive memory overhead of joint end-to-end training with large batch sizes, and (iii) the medical hallucinations induced by vanilla contrastive losses that ignore fine-grained anatomical laterality and negation modifiers. To address these challenges, we propose \textbf{SCALPEL}, a \textbf{S}emantic \textbf{C}ross-modal \textbf{A}lignment framework via \textbf{L}LM-\textbf{P}owered \textbf{E}ncoder \textbf{L}earning. First, Clinical Report Contrastive fine-tuning converts a generative LLM into an isotropic encoder via domain-specific clinical text adaptation. Second, an asymmetric alignment strategy leverages offline feature caching to enable efficient training. Critically, we formulate an Anatomy-Negation Aware Objective that explicitly penalizes mismatched image-text pairs involving laterality confusion or false negations. Extensive experiments across MIMIC-CXR, CheXpert, and IU X-Ray benchmarks demonstrate that SCALPEL achieves state-of-the-art performance in cross-modal retrieval, zero-shot disease classification and medical visual question answering.

\end{abstract}

\section{Introduction}

Medical multimodal representation learning has become a core technology in intelligent clinical decision support systems \cite{moor2023foundation}. Its objective is to map highly heterogeneous medical images and corresponding clinical text reports into a unified semantic representation space. Mature vision-language pretrained models form the foundation for various downstream tasks, including zero-shot disease classification and cross-modal text-image retrieval~\cite{zhang2022contrastive}. In modern clinical diagnostic workflows, treatment decisions rely heavily on fine-grained visual lesion features while integrating complete patient histories and complex pathological reasoning logic. Therefore, constructing a shared latent semantic space with strong representation capacity and discriminability is a necessary prerequisite for achieving precise automated clinical diagnosis~\cite{huang2021gloria}.

Although related research has made substantial progress, traditional medical vision-language representation learning frameworks \cite{wang2022medclip,boecking2022making} mostly adopt lightweight text encoders based on BERT \cite{devlin2019bert}. Shared representation spaces constructed upon such models face severe mismatches when processing coherent causal logic, multi-case comparison, and detailed pathological descriptions in real diagnostic workflows. Clinical reports are structurally rigorous professional medical texts filled with abundant medical terminology, diagnostic statements, and precise anatomical location descriptions. Meanwhile, the rapid development of pre-trained medical large language models \cite{wu2024pmcllama} has shown strong text understanding capabilities. Prior work \cite{huang2026llm2clip} attempted to extract textual features with frozen large language models, successfully unlocking long text understanding capabilities for CLIP architecture. Due to strict safety and complex semantics, directly applying LLMs degrades cross-modal matching to superficial lexical overlap, ignoring crucial diagnostic and spatial nuances in medical records.

\begin{figure*}[thb]
\centering
\vspace{-10pt}
\includegraphics[width=1.0\linewidth]{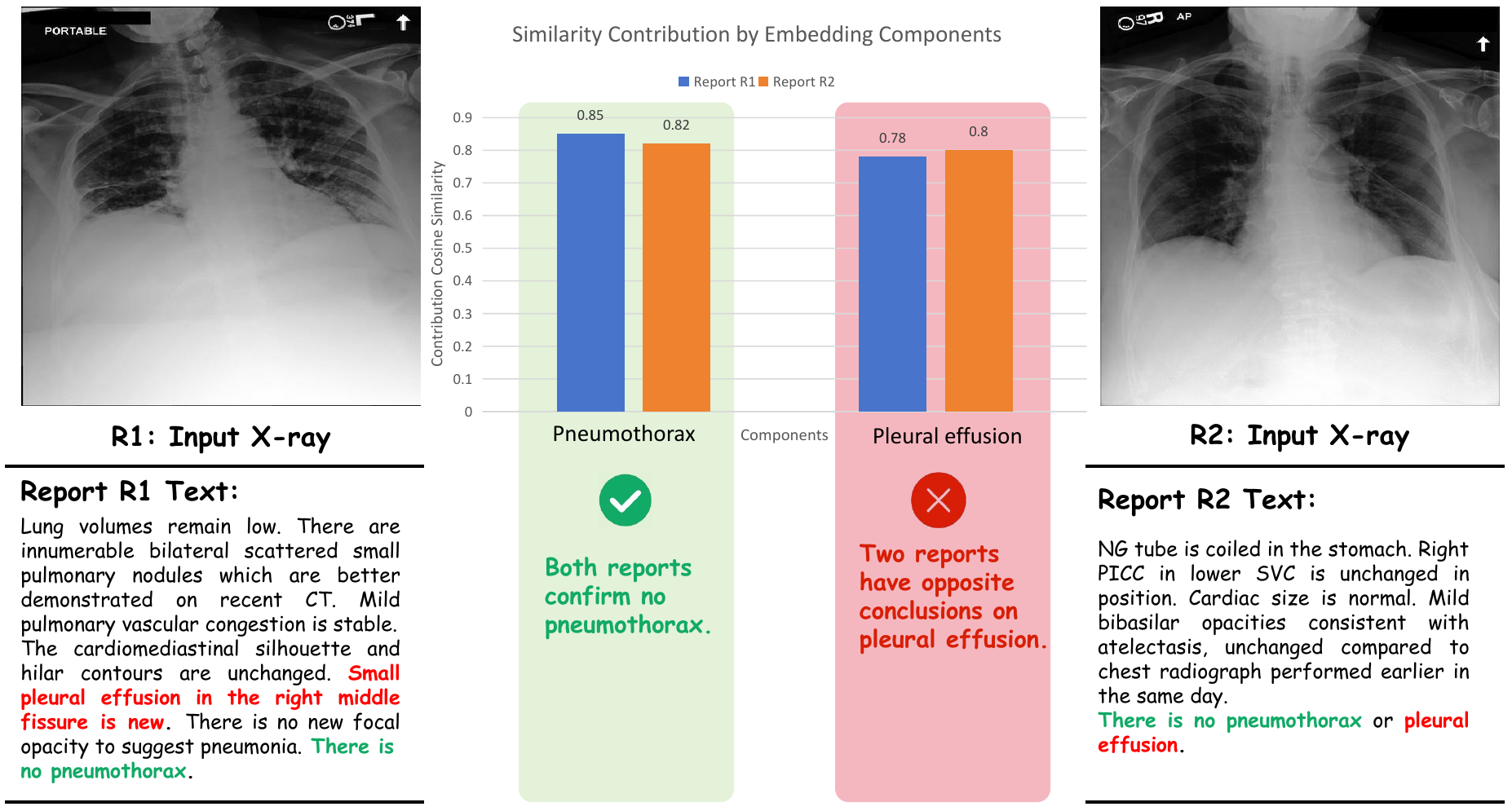}
\vspace{-10pt}
\caption{Vanilla medical contrastive learning models fail to assign differentiated scores to reports with opposite conclusions on MIMIC-CXR~\cite{johnson2019mimic}.}
\label{fig_teaser}
\vspace{-10pt}
\end{figure*}

Integrating generative medical large models with massive parameters into the visual feature alignment training pipeline introduces several bottlenecks. (i) Generative large language models optimize for autoregressive word-by-word prediction, causing the generated text embedding vectors to cluster, forming a narrow distribution interval \cite{gao2021simcse}. This anisotropic characteristic renders traditional contrastive learning evaluation metrics ineffective. As shown in \textbf{Fig}.~\ref{fig_teaser}, for two clinical reports with high lexical similarity but opposite diagnostic conclusions, the model struggles to distinguish their semantic differences and fails to assign well-differentiated similarity scores. (ii) End-to-end contrastive training of LLMs with high-resolution images and lengthy reports is computationally prohibitive. Consequently, existing methods compromise by truncating texts and compressing images, severely losing key diagnostic information. (iii) Conventional InfoNCE losses \cite{he2020momentum} prioritize frequent words over critical details like negations and laterality \cite{zhao2025clip}, causing severe medical hallucinations~\cite{wu2023medklip}.

To address these challenges, we propose \textbf{SCALPEL}, a \textbf{S}emantic \textbf{C}ross-modal \textbf{A}lignment framework via \textbf{L}LM-\textbf{P}owered \textbf{E}ncoder \textbf{L}earning for Medical Vision-Language Representation. SCALPEL follows a two-stage training paradigm. First, Clinical Report Contrastive (CRC) fine-tuning applies masked token prediction to radiology reports, enabling the LLM to model bidirectional clinical context. Second, an asymmetric alignment strategy freezes the LLM weights to ensure computationally efficient large-batch training. Finally, to mitigate the medical hallucinations induced by conventional contrastive losses, we formulate an Anatomy-Negation Aware Objective (ANAO). This objective introduces explicit penalty terms to penalize image-text pairs exhibiting conflicts in anatomical laterality and negation status. The main contributions of this paper are summarized as follows:

\begin{itemize}
\item We propose the SCALPEL, a cross-modal alignment architecture that fully exploits the rich semantic understanding capabilities of the pretrained medical large language model.
\item We design a CRC fine-tuning strategy, fine-tuning PMC-LLaMA on the MIMIC-CXR dataset to resolve the representation collapse problem inherent in generative large models.
\item We build an asymmetric training architecture integrated with offline feature caching and an anatomy-negation aware loss mechanism. This combination supports large-batch training while effectively avoiding medical semantic hallucinations.
\item Extensive experiments on MIMIC-CXR, IU X-Ray, and CheXpert demonstrate that SCALPEL achieves state-of-the-art performance in cross-modal retrieval and zero-shot disease classification, advancing the field of medical vision-language representation learning.
\end{itemize}

\section{Related Work}

\subsection{Conventional Medical Vision-Language Alignment}

Early medical Vision-Language Pre-training (VLP) methods adapt general-domain frameworks to clinical scenarios. Specifically, MedCLIP~\cite{wang2022medclip} decouples image--text pairs via soft semantic targets derived from medical ontologies, whereas BioViL~\cite{boecking2022making} introduces a radiology-specific BERT architecture. ConVIRT~\cite{zhang2022contrastive} and GLoRIA~\cite{huang2021gloria} explore local-global alignment between image patches and report sentences, and MedKLIP~\cite{wu2023medklip} uses external knowledge bases to augment supervision. However, these methods rely on lightweight, BERT-based text encoders with constrained context windows, limiting their capacity to model full clinical narratives. In contrast, SCALPEL utilizes the representational power of billion-parameter medical LLMs via an asymmetric design that maintains training efficiency.

\subsection{LLM-Enhanced Medical Vision-Language Models}

Integrating LLMs into vision-language learning has received growing attention. LLaVA-Med~\cite{li2023llava} and Med-Flamingo~\cite{moor2023med} adapt multimodal instruction-following frameworks from the general domain to medical applications. PMC-CLIP~\cite{lin2023pmc} and Med-PaLM M~\cite{tu2024towards} further investigate multimodal benchmarks enhanced by LLMs. In representation learning, LLM2CLIP~\cite{huang2026llm2clip} shows that caption-contrastive fine-tuning improves visual representations in the general domain. However, similar paradigms remain underexplored in the medical field. To bridge this gap, SCALPEL applies this methodology to healthcare and addresses three domain-specific challenges: report-length anisotropy, excessive memory consumption during joint training, and sensitivity to clinical negation.

\begin{figure*}[thb]
\centering
\includegraphics[width=1.0\linewidth]{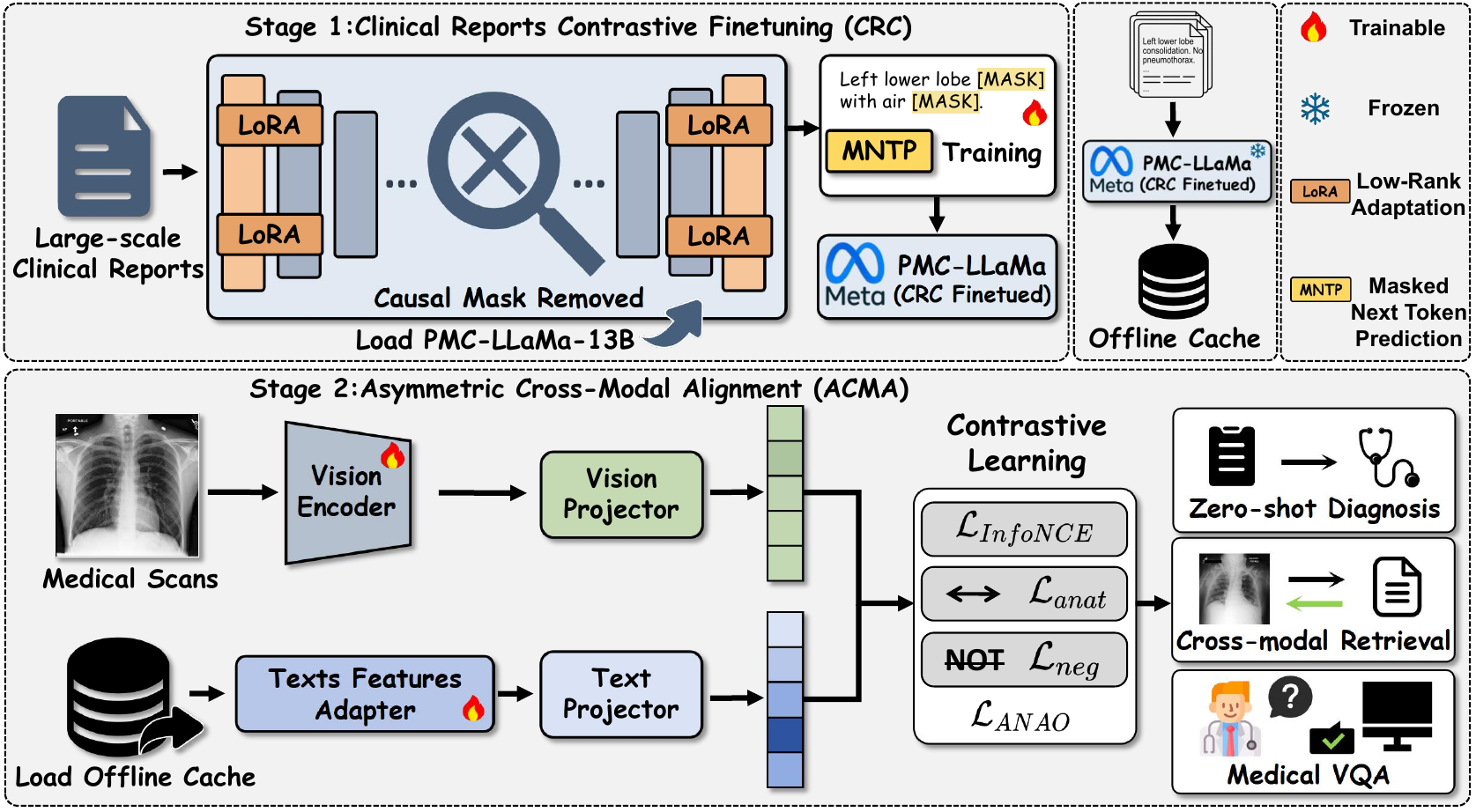}
\caption{Overview of the \textbf{SCALPEL} framework. Stage 1 (upper): CRC fine-tuning converts a generative medical LLM into a bidirectional text encoder via masked token prediction on radiology reports. Stage 2 (lower): the asymmetric alignment stage freezes the CRC-tuned LLM, pre-computes text features offline, and trains only the vision encoder and a lightweight cross-modal adapter under the ANAO loss.}
\label{fig_overview}
\end{figure*}

\section{Methodology}

\subsection{Overall Framework}
\label{sec:overall}

As illustrated in \textbf{Fig}.~\ref{fig_overview}, SCALPEL adopts a two-stage paradigm to map highly heterogeneous medical images and lengthy clinical reports into a unified, isotropic semantic space. Let $\mathcal{D} = \{(I_i, T_i)\}_{i=1}^{B}$ denote a training batch of $B$ chest radiograph images paired with their corresponding free-text radiology reports. We employ a medical large language model PMC-LLaMA-13B \cite{wu2024pmcllama} $\mathcal{E}_T$ to replace the original text encoder and a Vision Transformer DINOv2 ViT-B/16 \cite{oquab2023dinov2} $\mathcal{E}_V$ as the image encoder. Moreover, a lightweight multi-layer adapter $\mathcal{A}_{\text{proj}}$ projects the textual embeddings into a shared latent space of dimension $d$. 

The overarching objective of our framework builds upon the symmetric InfoNCE loss \cite{he2020momentum}, which maximizes the similarity of matched pairs while pushing apart $B-1$ unmatched pairs in the batch:
\begin{equation}
\mathcal{L}_{\text{InfoNCE}} = - \frac{1}{2B} \sum_{i=1}^{B} \left( \log \frac{\exp(\mathbf{S}_{i,i})}{\sum_{j=1}^{B} \exp(\mathbf{S}_{i,j})} + \log \frac{\exp(\mathbf{S}_{i,i})}{\sum_{j=1}^{B} \exp(\mathbf{S}_{j,i})} \right)
\label{eq:baseline_infonce}
\end{equation}
where $\mathbf{S}_{i,j} = \tau \cdot (\mathbf{v}_i^\top \tilde{\mathbf{t}}_j) / (\|\mathbf{v}_i\| \|\tilde{\mathbf{t}}_j\|)$ denotes the temperature-scaled cosine similarity between the visual embedding $\mathbf{v}_i$ and the projected text embedding $\tilde{\mathbf{t}}_j$.

\subsection{Stage 1: Clinical Report Contrastive (CRC) Fine-tuning}
\label{sec:crc}

To transform the LLM into a powerful discriminative encoder, we replace the causal attention mask with full bidirectional attention, enabling each token to attend to the entire clinical context. Given a radiology report $T$ tokenized into $\{x_1, \ldots, x_L\}$, we randomly mask a subset of positions $\mathcal{M}$. The MNTP objective is defined as:
\begin{equation}
\mathcal{L}_{\text{MNTP}} = -\frac{1}{|\mathcal{M}|} \sum_{m \in \mathcal{M}} \log P(x_m \mid x_{\setminus \mathcal{M}}; \Theta_{\text{LLM}} + \Delta\Theta_{\text{LoRA}})
\end{equation}
Simultaneously, we apply the SimCSE objective~\cite{gao2021simcse}. Each radiology report $T_i$ is passed through the encoder twice with independent dropout masks to construct positive pairs. This process can be denote as:
\begin{equation}
\mathcal{L}_{\text{SimCSE}} = -\frac{1}{B} \sum_{i=1}^{B} \log \frac{\exp\big( \cos(\mathbf{h}_i, \mathbf{h}_i^+) / \tau_s \big)}{\sum_{j=1}^{B} \exp\big( \cos(\mathbf{h}_i, \mathbf{h}_j^+) / \tau_s \big)}
\label{eq:simcse}
\end{equation}
where $\mathbf{h}_i = \text{MeanPool}(\mathcal{E}_T(T_i))$ and $\mathbf{h}_i^+ = \text{MeanPool}(\mathcal{E}_T(T_i^+))$ denote the pooled embeddings of the original and augmented reports, respectively, and $\tau_s$ is a text-only temperature parameter. The total Stage 1 objective is:
\begin{equation}
\mathcal{L}_{\text{CRC}} = \mathcal{L}_{\text{MNTP}} + \alpha \mathcal{L}_{\text{SimCSE}}
\end{equation}
where $\alpha$ balances the two terms. Only the low-rank adaptation \cite{hu2022lora} parameters ($\Delta\Theta_{\text{LoRA}}$) applied to the query, key, value, and output projection matrices receive gradient updates.

\subsection{Stage 2: Asymmetric Cross-modal Alignment}
\label{sec:stage2}
After CRC fine-tuning, we freeze LLM and pre-compute the dense text embeddings $\mathbf{t}_i = \mathcal{E}_T(T_i)$ for the entire training set. During cross-modal training, we load the cached text features $\mathbf{t}_i$ directly:
\begin{equation}
\mathbf{v}_i = \mathcal{E}_V(I_i), \quad \tilde{\mathbf{t}}_i = \mathcal{A}_{\text{proj}}(\mathbf{t}_i)
\end{equation}
where $\mathcal{A}_{\text{proj}}$ is a lightweight 4-layer MLP that maps the 5120-dimensional LLM embeddings to the $d$-dimensional shared space.

\noindent\textbf{Anatomy-Negation Aware Objective (ANAO)}
In radiology, two reports may share over $90\%$ lexical overlap yet carry opposite diagnostic implications. Vanilla InfoNCE is influenced by high-frequency shared tokens and fails to penalize these critical semantic mismatches.

We augment the standard InfoNCE loss with two additive penalty terms that operate on pre-extracted clinical metadata. For each report $T_i$, we use clinical NER pipeline \cite{velamala2025medspacyv,su2025large} to extract two structured labels: an anatomical label encoding locations ($0$: left\_upper, $1$: left\_lower, $2$: right\_upper, $3$: right\_lower, $4$: bilateral, $5$: unspecified), and a negation vector $\mathbf{n}_i \in \{0,1\}^{K}$ where $K$ denotes the numbers of radiological findings (e.g., pneumothorax, consolidation, cardiomegaly). The ANAO objective is defined as:

\begin{equation}
\mathcal{L}_{\text{ANAO}} = \mathcal{L}_{\text{InfoNCE}} + \lambda_a \mathcal{L}_{\text{anat}} + \lambda_n \mathcal{L}_{\text{neg}}
\label{eq:anao}
\end{equation}
where $\mathcal{L}_{\text{InfoNCE}}$ is defined in Eq.~\ref{eq:baseline_infonce}, and $\lambda_a$, $\lambda_n$ are hyperparameters.

\textbf{Anatomy penalty $\mathcal{L}_{\text{anat}}$.} This term penalizes high image--text similarity when the two reports describe findings at different anatomical locations. Concretely, consider a batch where report $i$ describes a left-upper-lobe consolidation ($a_i{=}0$) and report $j$ describes a right-upper-lobe nodule ($a_j{=}2$). If the model spuriously assigns high similarity $\mathbf{S}_{ij}$ to this mismatched pair, $\mathcal{L}_{\text{anat}}$ contributes a positive penalty proportional to the similarity magnitude, driving the model to separate left-sided from right-sided representations:

\begin{equation}
\mathcal{L}_{\text{anat}} = \frac{1}{|\mathcal{M}_a|} \sum_{(i,j) \in \mathcal{M}_a} \max(0, \mathbf{S}_{ij}), \quad \mathcal{M}_a = \{(i,j) \mid a_i \neq a_j \}
\label{eq:anat}
\end{equation}

The $\max(0, \cdot)$ operator ensures only pairs with high cosine similarity receive penalties.

\textbf{Negation penalty $\mathcal{L}_{\text{neg}}$.} For each pair $(i,j)$, we compute the normalized Hamming distance between their negation vectors as the fraction of findings whose negation status disagrees. This distance serves as a penalty weight: pairs differing on many negation dimensions receive stronger penalization if their similarity $\mathbf{S}_{ij}$ is spuriously high:

\begin{equation}
\mathcal{L}_{\text{neg}} = \frac{1}{B^2} \sum_{i,j} \left( \frac{1}{K} \sum_{k=1}^{K} \mathbb{I}[\mathbf{n}_{i,k} \neq \mathbf{n}_{j,k}] \right) \cdot \max(0, \mathbf{S}_{ij})
\label{eq:neg}
\end{equation}

For example, if report $i$ contains ``no pneumothorax'' ($\mathbf{n}_{i,\text{pneumo}}{=}0$) while report $j$ describes a pneumothorax-positive image ($\mathbf{n}_{j,\text{pneumo}}{=}1$), the negation dimension contributes to the penalty. Reports that agree on most negation states produce a small penalty weight even at high similarity, preserving legitimate matches between reports that differ on only minor findings.

\section{Experiments}

\subsection{Datasets and Evaluation Metrics}

\noindent\textbf{Data Preparation.} We evaluate our framework across five medical benchmarks to comprehensively demonstrate its capabilities. For cross-modal retrieval, we utilize \textbf{MIMIC-CXR}~\cite{johnson2019mimic}, the largest publicly available chest radiograph dataset, comprising 377,110 images from 227,835 studies with corresponding de-identified radiology reports. For downstream zero-shot disease classification, we employ \textbf{IU X-Ray}~\cite{demner2016preparing} and \textbf{CheXpert}~\cite{irvin2019chexpert} to evaluate the models' clinical generalization capabilities without any task-specific fine-tuning. IU X-Ray contains 7,470 chest radiographs paired with 3,955 radiology reports. CheXpert consists of 224,316 chest radiographs collected from 65,240 patients, annotated across 14 distinct radiological observations. Following the zero-shot evaluation protocol in~\cite{elallaf2026medprobclip}, we cast disease classification as an image-text matching task, where the prediction is assigned based on the maximum cosine similarity between the input visual embedding and standardized categorical prompts. For medical visual question answering (VQA), we evaluate on \textbf{VQA-RAD}~\cite{lau2018VQARADdataset} and \textbf{SLAKE}~\cite{liu2021slake}. VQA-RAD contains 315 radiology images and 3,515 clinician-annotated question-answer pairs. SLAKE is a knowledge-enhanced benchmark comprising 642 radiology slices across multiple modalities, including X-rays, computerized tomography, and magnetic resonance imaging, paired with over 7,000 question-answer pairs.

\noindent\textbf{Evaluation Metrics.} To rigorously assess the proposed framework, distinct evaluation metrics are adopted tailored to each specific downstream task. 
For cross-modal retrieval on MIMIC-CXR, we report Recall@$k$ ($k \in \{1, 5, 10\}$) for both image-to-text (I2T) and text-to-image (T2I) retrieval tasks. 
For zero-shot disease classification on IU X-Ray and CheXpert, we compute the Accuracy (ACC) and the macro F1-score to measure diagnostic accuracy under highly imbalanced clinical condition distributions. 
For medical visual question answering on VQA-RAD and SLAKE, we follow the prior work \cite{MMedPO_ICML25}. For open-ended questions, we report recall (Open); for closed-ended questions, accuracy (Closed).

\subsection{Implementation Details}

\noindent\textbf{CRC Fine-tuning.} We initialize the text encoder from PMC-LLaMA-13B~\cite{wu2024pmcllama} and apply LoRA~\cite{hu2022lora} with rank $r{=}8$, $\alpha{=}16$, and dropout ${=}0.05$ to all attention projection matrices. Bidirectional attention is enabled by zeroing the causal mask. The model is trained with the MNTP objective for 10,000 steps on the MIMIC-CXR training sets, using a batch size of 64, a maximum sequence length of 512 tokens, and the AdamW optimizer with a cosine learning rate schedule (peak LR $2{\times}10^{-4}$).

\noindent\textbf{Cross-modal Alignment.} Training uses the ANAO loss with $\lambda_a=0.1$, $\lambda_n=0.15$, batch size 128, learning rate $10^{-4}$ for both encoders, weight decay 0.01, linear warmup over 2,000 steps followed by cosine decay, and automatic mixed precision (bf16). For the full SCALPEL configuration, we replace the text encoder with a frozen CRC-tuned PMC-LLaMA-13B \cite{wu2024pmcllama} and the vision encoder with DINOv2 ViT-B/16 \cite{oquab2023dinov2}.
\begin{table}[ht]
\caption{Cross-modal retrieval performance on MIMIC-CXR \cite{johnson2019mimic}. \colorbox{gray!10}{\textit{+SCALPEL}} indicates replacing the original text encoder and contrastive loss with our PMC-LLaMA-based encoder and ANAO objective while preserving the original vision encoder. \colorbox{gray!25}{\textbf{SCALPEL*}} denotes our fully instantiated model.}
\label{tab:retrieval_sources}
\vspace{-20pt}
\begin{center}
\resizebox{\columnwidth}{!}{
\begin{tabular}{cccccccc}
\toprule
\multirow{2}{*}{\textbf{Methods}} & \multirow{2}{*}{\textbf{Sources}} & \multicolumn{3}{c}{\textbf{I2T}} & \multicolumn{3}{c}{\textbf{T2I}} \\
\cmidrule(lr){3-5} \cmidrule(lr){6-8}
 & & \textbf{R@1} & \textbf{R@5} & \textbf{R@10} & \textbf{R@1} & \textbf{R@5} & \textbf{R@10} \\
\midrule
Med-CLIP \cite{wang2022medclip}& EMNLP'22       & 14.24 & 28.72 & 41.07 & 17.45 & 32.49 & 41.25 \\
\cellcolor{gray!10}{\;\;\textit{+SCALPEL}} &  \cellcolor{gray!10}               & \cellcolor{gray!10}{16.52} & \cellcolor{gray!10}{31.54} & \cellcolor{gray!10}{45.92} & \cellcolor{gray!10}{18.47} & \cellcolor{gray!10}{34.58} & \cellcolor{gray!10}{43.97} \\
CXR-CLIP \cite{you2023cxr}& MICCAI'23      & 17.08 & 41.43 & 52.42 & 16.79 & 42.92 & 54.45 \\
\cellcolor{gray!10}{\;\;\textit{+SCALPEL}} & \cellcolor{gray!10}                & \cellcolor{gray!10}{18.43} & \cellcolor{gray!10}{43.27} & \cellcolor{gray!10}{53.45} & \cellcolor{gray!10}{18.51} & \cellcolor{gray!10}{43.27} & \cellcolor{gray!10}{54.98} \\
BioViL-T \cite{Bannur_2023_CVPR}& CVPR'23        & 20.12 & 39.48 & 52.37 & 18.72 & 44.83 & 52.63 \\
\cellcolor{gray!10}{\;\;\textit{+SCALPEL}} &  \cellcolor{gray!10}               & \cellcolor{gray!10}{20.73} & \cellcolor{gray!10}{40.22} & \cellcolor{gray!10}{52.98} & \cellcolor{gray!10}{19.02} & \cellcolor{gray!10}{42.68} & \cellcolor{gray!10}{52.95} \\
BiomedCLIP \cite{BiomedCLIP}& NEJMAI'25    & 19.74 & 39.43 & 50.73 & 18.53 & 41.84 & 50.07 \\
\cellcolor{gray!10}{\;\;\textit{+SCALPEL}} & \cellcolor{gray!10}                & \cellcolor{gray!10}{21.06} & \cellcolor{gray!10}{42.51} & \cellcolor{gray!10}{52.83} & \cellcolor{gray!10}{\underline{19.07}} & \cellcolor{gray!10}{43.21} & \cellcolor{gray!10}{54.43} \\
MedProbCLIP \cite{elallaf2026medprobclip}& WACV'26     & 20.34 & 42.71 & 53.48 & 18.73 & 46.48 & 55.74 \\
\cellcolor{gray!10}{\;\;\textit{+SCALPEL}} & \cellcolor{gray!10}              & \cellcolor{gray!10}{\underline{21.89}} & \cellcolor{gray!10}{\underline{43.96}} & \cellcolor{gray!10}{\underline{55.27}} & \cellcolor{gray!10}{19.03} & \cellcolor{gray!10}{\underline{46.87}} & \cellcolor{gray!10}{\underline{56.21}} \\
PCME++ \cite{chun2024improved}& ICLR'24          & 18.73 & 32.77 & 47.75 & 16.54 & 39.45 & 48.53 \\
MMedPO \cite{MMedPO_ICML25}& ICML'25          & 21.46 & 33.48 & 52.44 & 17.82 & 45.26 & 51.48 \\
\midrule
\cellcolor{gray!25}{\textbf{SCALPEL*}} & \cellcolor{gray!25}{\textbf{Ours}}      & \cellcolor{gray!25}\textbf{23.17} & \cellcolor{gray!25}\textbf{45.26} & \cellcolor{gray!25}\textbf{54.87} & \cellcolor{gray!25}\textbf{19.38} & \cellcolor{gray!25}\textbf{47.13} & \cellcolor{gray!25}\textbf{56.77} \\
\bottomrule
\end{tabular}
} 
\vspace{-20pt}
\end{center}
\end{table}
\subsection{Main Results}

\noindent\textbf{Cross-modal Retrieval.}
Table~\ref{tab:retrieval_sources} presents retrieval results on MIMIC-CXR \cite{johnson2019mimic}. SCALPEL acts as an adaptable module that consistently improves existing baselines. For example, integrating SCALPEL into Med-CLIP \cite{wang2022medclip} increases I2T R@1 by 2.28\%  and T2I R@1 by 1.02\% . Our fully instantiated model achieves the highest performance, reaching 23.17\% on I2T R@1 and 19.38\% on T2I R@1.

We observe two specific metric decreases due to architectural trade-offs. First, BioViL-T+SCALPEL's T2I R@5 drops from 44.83\% to 42.68\%. BioViL-T \cite{Bannur_2023_CVPR} uses a ResNet-50 backbone for localized token alignment. The ANAO objective imposes strict anatomy and negation constraints, prioritizing exact top-1 matches (T2I R@1 improves to 19.02\%) over looser semantic associations at R@5. Second, SCALPEL* (54.87\%) trails MedProbCLIP+SCALPEL (55.27\%) on I2T R@10. MedProbCLIP uses probabilistic embeddings to model clinical ambiguity as spatial distributions. At higher recall ranks, this distribution-based approach captures broad semantic variations more effectively than the rigid point-to-point alignment enforced within our isotropic semantic space.

\begin{table}[htbp]
\caption{Downstream generalization performance across various medical benchmarks. \colorbox{gray!10}{\textit{+SCALPEL}} indicates replacing the original text encoder with our LLM-driven strategy. \colorbox{gray!25}{\textbf{SCALPEL*}} denotes our fully instantiated model.}
\label{tab:downstream_tasks}
\vspace{-20pt}
\begin{center}
\resizebox{\textwidth}{!}{
\begin{tabular}{cccccccccc}
\toprule
\multirow{2}{*}{\textbf{Methods}} & \multirow{2}{*}{\textbf{Sources}} & \multicolumn{2}{c}{\textbf{IU X-Ray}\cite{demner2016preparing}} & \multicolumn{2}{c}{\textbf{CheXpert}\cite{irvin2019chexpert}} & \multicolumn{2}{c}{\textbf{SLAKE}\cite{liu2021slake}} & \multicolumn{2}{c}{\textbf{VQA-RAD}\cite{lau2018VQARADdataset}} \\
\cmidrule(lr){3-4} \cmidrule(lr){5-6} \cmidrule(lr){7-8} \cmidrule(lr){9-10}
 & & ACC & F1 & ACC & F1 & Open & Closed & Open & Closed \\
\midrule

Med-CLIP \cite{wang2022medclip} & EMNLP'22 & 0.632 & 0.506 & 0.538 & 0.507 & 43.17 & 59.84 & 29.48 & 58.43 \\
\cellcolor{gray!10}{\;\;\textit{+SCALPEL}} & \cellcolor{gray!10} & \cellcolor{gray!10}{0.655} & \cellcolor{gray!10}{0.541} & \cellcolor{gray!10}{0.549} & \cellcolor{gray!10}{0.523} & \cellcolor{gray!10}{44.32} & \cellcolor{gray!10}{60.03} & \cellcolor{gray!10}{32.14} & \cellcolor{gray!10}{61.47} \\

CXR-CLIP \cite{you2023cxr}& MICCAI'23 & 0.652 & 0.547 & 0.543 & 0.515 & 42.84 & 60.32 & 29.33 & 57.4 \\
\cellcolor{gray!10}{\;\;\textit{+SCALPEL}} & \cellcolor{gray!10} & \cellcolor{gray!10}{0.671} & \cellcolor{gray!10}{0.583} & \cellcolor{gray!10}{0.564} & \cellcolor{gray!10}{0.572} & \cellcolor{gray!10}{45.01} & \cellcolor{gray!10}{58.49} & \cellcolor{gray!10}{30.18} & \cellcolor{gray!10}{58.26} \\

BioViL-T \cite{Bannur_2023_CVPR}& CVPR'23 & 0.683 & 0.627 & 0.557 & 0.612 & 51.43 & 61.37 & 33.47 & 52.48 \\
\cellcolor{gray!10}{\;\;\textit{+SCALPEL}} & \cellcolor{gray!10} & \cellcolor{gray!10}{0.677} & \cellcolor{gray!10}{\underline{0.584}} & \cellcolor{gray!10}{0.542} & \cellcolor{gray!10}{0.619} & \cellcolor{gray!10}{51.74} & \cellcolor{gray!10}{63.48} & \cellcolor{gray!10}{\underline{35.75}} & \cellcolor{gray!10}{53.74} \\

BiomedCLIP \cite{BiomedCLIP}& NEJMAI'25 & 0.545 & 0.559 & 0.486 & 0.524 & 47.23 & 62.81 & 32.16 & 59.45 \\
\cellcolor{gray!10}{\;\;\textit{+SCALPEL}} & \cellcolor{gray!10} & \cellcolor{gray!10}{0.576} & \cellcolor{gray!10}{0.564} & \cellcolor{gray!10}{0.508} & \cellcolor{gray!10}{0.593} & \cellcolor{gray!10}{47.49} & \cellcolor{gray!10}{63.44} & \cellcolor{gray!10}{34.43} & \cellcolor{gray!10}{62.44} \\

MedProbCLIP \cite{elallaf2026medprobclip}& WACV'26 & 0.648 & 0.455 & 0.526 & 0.614 & 51.56 & 63.43 & 32.13 & 62.83 \\
\cellcolor{gray!10}{\;\;\textit{+SCALPEL}} & \cellcolor{gray!10} & \cellcolor{gray!10}{\underline{0.692}} & \cellcolor{gray!10}{0.467} & \cellcolor{gray!10}{0.532} & \cellcolor{gray!10}{\underline{0.623}} & \cellcolor{gray!10}{\underline{52.07}} & \cellcolor{gray!10}{64.75} & \cellcolor{gray!10}{33.98} & \cellcolor{gray!10}{63.41} \\

PCME++ \cite{chun2024improved}& ICLR'24 & 0.541 & 0.524 & 0.483 & 0.442 & 48.52 & 60.17 & 32.52 & 61.95 \\

MMedPO \cite{MMedPO_ICML25}& ICML'25 & 0.587 & 0.54 & \underline{0.575} & 0.608 & 50.75 & \underline{65.23} & 35.78 & \underline{65.79} \\

\midrule
\cellcolor{gray!25}{\textbf{SCALPEL*}} & \cellcolor{gray!25}{\textbf{Ours}} & \cellcolor{gray!25}{\textbf{0.694}} & \cellcolor{gray!25}{\textbf{0.634}} & \cellcolor{gray!25}{\textbf{0.581}} & \cellcolor{gray!25}{\textbf{0.637}} & \cellcolor{gray!25}{\textbf{53.86}} & \cellcolor{gray!25}{\textbf{66.74}} & \cellcolor{gray!25}{\textbf{36.39}} & \cellcolor{gray!25}{\textbf{66.54}} \\

\bottomrule
\end{tabular}
} 
\vspace{-20pt}
\end{center}
\end{table}

\noindent\textbf{Downstream Task Validation}
Table~\ref{tab:downstream_tasks} evaluates zero-shot transfer to disease classification and VQA. Our strategy consistently enhances baseline generalization by constructing an isotropic semantic space and enforcing anatomy-negation constraints. Quantitatively, SCALPEL* achieves the highest performance, surpassing MedProbCLIP by 4.6\% in IU X-Ray ACC and MMedPO by 3.11\% in SLAKE Open VQA. As a plug-and-play module, +SCALPEL improves Med-CLIP's CheXpert ACC by 1.1\% and SLAKE Open by +1.15\%. However, BioViL-T+SCALPEL exhibits a drop of 0.043 in IU X-Ray F1. This decrease occurs because the ANAO objective imposes strict penalty constraints on anatomical orientation and false negations. BioViL-T relies on a lightweight text encoder with a constrained context window. Without the bidirectional context modeling provided by our CRC fine-tuning on large language models, this limited encoder fails to resolve the complex semantic boundaries demanded by the ANAO loss, resulting in reduced matching accuracy.

\subsection{Ablation Study and Efficiency Analysis}
\begin{figure*}[thb]
\centering
\vspace{-10pt}
\includegraphics[width=1.0\linewidth]{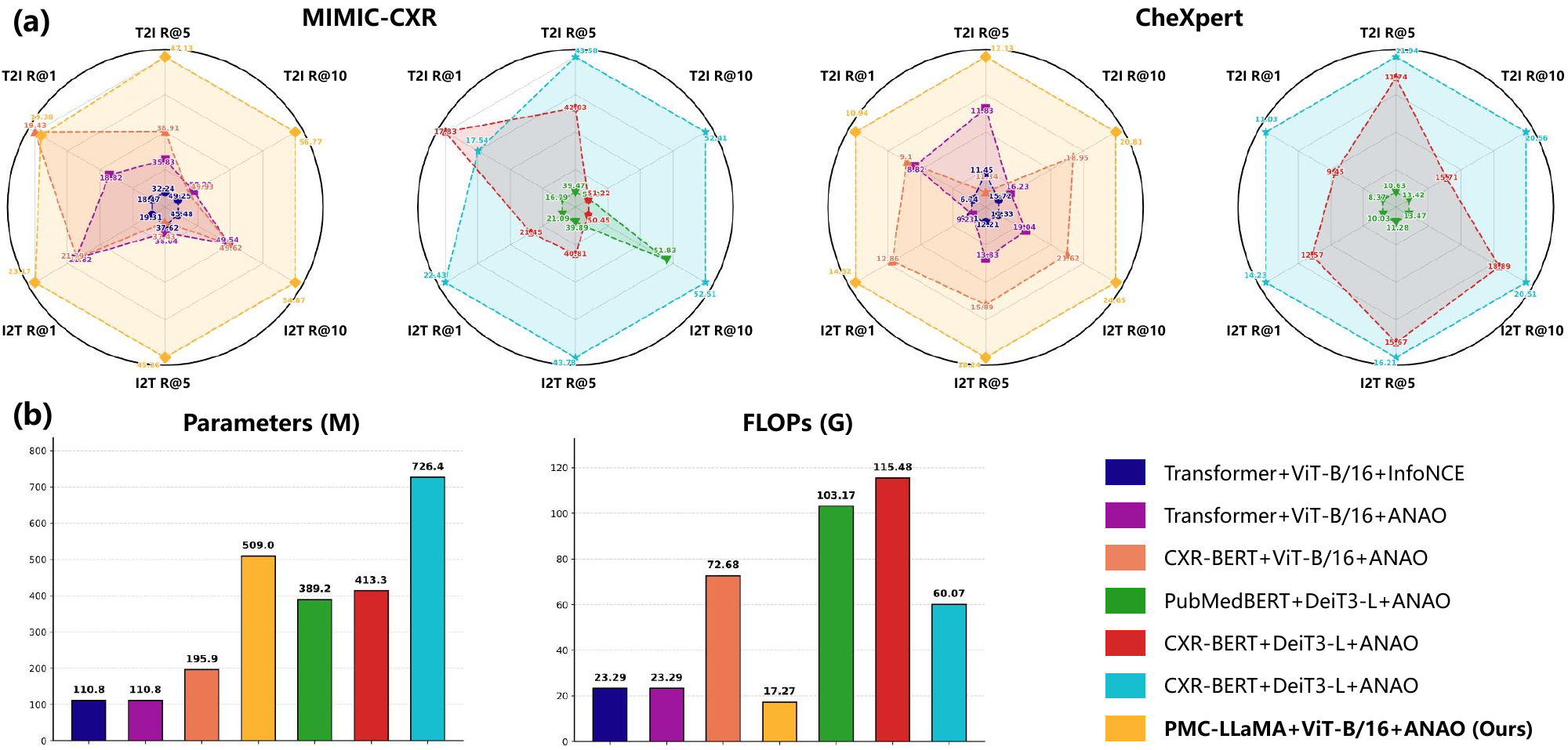}
\caption{Ablation study and efficiency analysis on MIMIC-CXR~\cite{johnson2019mimic} and CheXpert\cite{irvin2019chexpert}.}
\label{fig_ablation}
\vspace{-10pt}
\end{figure*}

\textbf{Fig}.~\ref{fig_ablation} (a) illustrates the ablation results across MIMIC-CXR~\cite{johnson2019mimic} and CheXpert~\cite{irvin2019chexpert} datasets, demonstrating the superior performance of our model.

On MIMIC-CXR~\cite{johnson2019mimic} dataset, the introduction of the ANAO loss to the vanilla InfoNCE baseline yields consistent gains, improving I2T R@1 from 19.31\% to 21.82\%. This confirms the efficacy of the anatomy-negation penalty in mitigating semantic hallucinations. Furthermore, upgrading the text encoder from CXR-BERT\cite{boecking2022making} to our CRC-tuned PMC-LLaMA delivers a significant boost, driving I2T R@1 to 23.17\% and T2I R@1 to 19.38\%. This substantial improvement highlights the critical role of bidirectional context modeling—achieved via CRC fine-tuning on the generative LLM—in capturing long-range clinical dependencies better than lightweight BERT models. The generalization capability of this architecture is strongly validated on the CheXpert dataset~\cite{irvin2019chexpert}. The SCALPEL configuration achieves the highest performance across all metrics, with I2T R@1 reaching 14.52\% and T2I R@1 reaching 10.94\%, outperforming the strongest baseline (12.86\% and 9.10\%). Although DeiT3-L possesses greater capacity, the ViT-B/16 backbone combined with our robust text encoder and ANAO objective proves to be the most optimal and efficient alignment strategy. These structural advantages justify our selection of the SCALPEL components, effectively balancing high-precision cross-modal matching with computational feasibility.

Efficiency tests in \textbf{Fig}.~\ref{fig_ablation} (b) further validate the structural superiority of the chosen SCALPEL configuration. The parameter footprint of our model reaches 509M, which originates from the wide multi-layer perceptron required to project the high-dimensional LLM embeddings into the shared latent space. Despite this larger parameter size, SCALPEL dramatically minimizes the actual computational burden. Specifically, it requires only 17.27G FLOPs, representing a substantial 76\% reduction compared to the 72.68G FLOPs consumed by the CXR-BERT baseline. This high computational efficiency is a direct result of our asymmetric design, which leverages offline feature caching to bypass the heavy dynamic sequence computations of standard text encoders. Furthermore, replacing ViT-B/16 with a heavier DeiT3-L backbone inflates the parameters to 726.4M and FLOPs to 60.07G, yet paradoxically leads to inferior retrieval performance (e.g., I2T R@1 drops from 23.17\% to 22.43\%). This demonstrates that blindly scaling up the vision encoder does not guarantee better cross-modal alignment in our medical setting, likely due to overfitting on the limited pre-training data scale. Consequently, the ViT-B/16 backbone strikes the optimal balance, successfully incorporating rich clinical priors from massive language models while preserving a highly lightweight and deployable operational cost.

\begin{table}[ht]
\caption{Hyperparameter sensitivity study on MIMIC-CXR \cite{johnson2019mimic}. Best and second best results are highlighted in \textbf{bold} and \underline{underlined}, respectively.}
\label{tab:hyperparams}
\vspace{-10pt}
\begin{center}
\begin{tabular}{ccc ccc ccc}
\toprule
\multicolumn{3}{c}{\textbf{Hyperparameter}} & \multicolumn{3}{c}{\textbf{I2T}} & \multicolumn{3}{c}{\textbf{T2I}} \\
\cmidrule(lr){1-3} \cmidrule(lr){4-6} \cmidrule(lr){7-9}
$\alpha$ & $\lambda_a$ & $\lambda_n$ & \textbf{R@1} & \textbf{R@5} & \textbf{R@10} & \textbf{R@1} & \textbf{R@5} & \textbf{R@10} \\
\midrule
0.1 & 0.1  & 0.15 & 21.50 & 40.12 & 51.20 & 16.85 & 43.20 & 49.50 \\
0.3 & 0.1  & 0.15 & 22.95 & 42.15 & \underline{53.80} & 18.12 & 45.60 & 53.15 \\
\textbf{0.5} & 0.1 & 0.15 & \textbf{23.60} & \textbf{45.26} & \textbf{54.87} & \textbf{19.38} & \textbf{47.13} & \textbf{56.77} \\
0.7 & 0.1  & 0.15 & \underline{23.54} & \underline{43.05} & 52.90 & \underline{18.70} & \underline{46.25} & \underline{54.80} \\
0.9 & 0.1  & 0.15 & 22.40 & 41.80 & 51.65 & 17.40 & 44.90 & 52.10 \\
\midrule
0.5 & 0.05 & 0.15 & 22.73 & 40.94 & 53.14 & 17.62 & 45.74 & \underline{52.38} \\
0.5 & \textbf{0.1} & 0.15 & \textbf{23.17} & \textbf{45.26} & \textbf{54.87} & \textbf{19.38} & \textbf{47.13} & \textbf{56.77} \\
0.5 & 0.15 & 0.15 & 23.21 & \underline{41.32} & 52.73 & 16.43 & 43.18 & 48.92 \\
0.5 & 0.2  & 0.15 & \underline{24.02} & 39.44 & \underline{53.62} & \underline{19.19} & \underline{46.31} & 51.99 \\
0.5 & 0.25 & 0.15 & 23.75 & 35.48 & 49.41 & 17.51 & 42.89 & 45.97 \\
\midrule
0.5 & 0.1  & 0.05 & 21.87 & 42.79 & \underline{52.43} & \underline{18.43} & \underline{46.85} & 51.42 \\
0.5 & 0.1  & 0.1  & 22.51 & 42.40 & 51.55 & 16.27 & 44.83 & 53.48 \\
0.5 & 0.1  & \textbf{0.15} & \textbf{23.17} & \textbf{45.26} & \textbf{54.87} & \textbf{19.38} & \textbf{47.13} & \textbf{56.77} \\
0.5 & 0.1  & 0.2  & \underline{22.94} & \underline{43.14} & 50.93 & 17.21 & 45.49 & \underline{53.92} \\
0.5 & 0.1  & 0.25 & 22.15 & 42.93 & 49.74 & 17.09 & 44.98 & 49.84 \\
\bottomrule
\end{tabular}
\vspace{-20pt}
\end{center}
\end{table}

\subsection{Hyperparameter Sensitivity Analysis}
\label{sec:hyperparams}
Table~\ref{tab:hyperparams} provides detailed analysis regarding hyperparameter selection, demonstrating that SCALPEL achieves optimal performance at $\alpha=0.5$, $\lambda_a=0.1$, and $\lambda_n=0.15$. Quantitatively, setting $\alpha$ to 0.5 achieves the highest I2T R@1 of 23.60\%, outperforming $\alpha=0.1$ (21.50\%) by 9.77\%. This confirms that a balanced $\alpha$ optimally breaks the LLM's anisotropic collapse via the SimCSE loss without overwhelming the bidirectional context established by the MNTP objective. Regarding the ANAO objective, $\lambda_a=0.1$ and $\lambda_n=0.15$ establish the best result. Reducing these weights ($\lambda_a=0.05$ or $\lambda_n=0.05$) degrades I2T R@1 to 22.73\% and 21.87\%, respectively, failing to sufficiently penalize anatomical laterality confusion and negation blindness. Conversely, excessively high penalties ($\lambda_a=0.25$ or $\lambda_n=0.25$) sharply decrease T2I R@1 to 17.51\% and 17.09\%, as they over-constrain the representation and disrupt the temperature-scaled gradient dynamics of the primary InfoNCE loss. These precise parameter choices ensure SCALPEL strictly enforces fine-grained clinical constraints without compressing the broad semantic matching capacity of the shared isotropic space.

\section{Conclusion}

\label{sec:conclusion}
We presented SCALPEL, an LLM-powered encoder learning framework that resolves the representational and computational bottlenecks of medical vision-language alignment. Functioning as an adaptable modular toolkit, it decomposes the alignment process into CRC fine-tuning, asymmetric feature caching, and the ANAO objective, allowing independent integration into existing VLP pipelines. Extensive evaluations across cross-modal retrieval, zero-shot classification, and VQA benchmarks confirm its superior performance and robust clinical generalization. Despite these advances, SCALPEL's reliance on external NLP pipelines for ANAO metadata extraction may introduce cascading errors, and freezing the LLM prevents deep, early-stage cross-modal attention. Future work will explore end-to-end LLM-driven extraction and parameter-efficient fine-tuning to enable deeper multimodal fusion while maintaining training efficiency.

\bibliographystyle{splncs04}
\bibliography{main}

\begin{thebibliography}{10}
\providecommand{\url}[1]{\texttt{#1}}
\providecommand{\urlprefix}{URL }
\providecommand{\doi}[1]{https://doi.org/#1}

\bibitem{Bannur_2023_CVPR}
Bannur, S., Hyland, S., Liu, Q., P\'erez-Garc{\'\i}a, F., Ilse, M., Castro, D.C., Boecking, B., Sharma, H., Bouzid, K., Thieme, A., Schwaighofer, A., Wetscherek, M., Lungren, M.P., Nori, A., Alvarez-Valle, J., Oktay, O.: Learning to exploit temporal structure for biomedical vision-language processing. In: Proceedings of the IEEE/CVF Conference on Computer Vision and Pattern Recognition (CVPR). pp. 15016--15027 (June 2023)

\bibitem{boecking2022making}
Boecking, B., Usuyama, N., Bannur, S., Castro, D.C., Schwaighofer, A., Hyland, S., Wetscherek, M., Naumann, T., Nori, A., Alvarez-Valle, J., et~al.: Making the most of text semantics to improve biomedical vision--language processing. In: European conference on computer vision. pp. 1--21. Springer (2022)

\bibitem{chun2024improved}
Chun, S.: Improved probabilistic image-text representations. In: International Conference on Learning Representations. vol.~2024, pp. 39787--39815 (2024)

\bibitem{demner2016preparing}
Demner-Fushman, D., Kohli, M.D., Rosenman, M.B., Shooshan, S.E., Rodriguez, L., Antani, S., Thoma, G.R., McDonald, C.J.: Preparing a collection of radiology examinations for distribution and retrieval. Journal of the American Medical Informatics Association  \textbf{23}(2),  304--310 (2016)

\bibitem{devlin2019bert}
Devlin, J., Chang, M.W., Lee, K., Toutanova, K.: {BERT}: Pre-training of deep bidirectional transformers for language understanding. In: Burstein, J., Doran, C., Solorio, T. (eds.) Proceedings of the 2019 Conference of the North {A}merican Chapter of the Association for Computational Linguistics: Human Language Technologies, Volume 1 (Long and Short Papers). pp. 4171--4186. Association for Computational Linguistics, Minneapolis, Minnesota (Jun 2019)

\bibitem{elallaf2026medprobclip}
Elallaf, A., Zhang, Y., Masupalli, Y., Yang, J., Lee, Y., Cao, Z., Liang, G.: {MedProbCLIP}: Probabilistic adaptation of vision-language foundation model for reliable radiograph-report retrieval. In: Proceedings of the IEEE/CVF Winter Conference on Applications of Computer Vision. pp. 1--10 (2026)

\bibitem{gao2021simcse}
Gao, T., Yao, X., Chen, D.: {SimCSE}: Simple contrastive learning of sentence embeddings. In: Proceedings of the 2021 conference on empirical methods in natural language processing. pp. 6894--6910 (2021)

\bibitem{he2020momentum}
He, K., Fan, H., Wu, Y., Xie, S., Girshick, R.: Momentum contrast for unsupervised visual representation learning. In: Proceedings of the IEEE/CVF conference on computer vision and pattern recognition. pp. 9729--9738 (2020)

\bibitem{hu2022lora}
Hu, E.J., Shen, Y., Wallis, P., Allen-Zhu, Z., Li, Y., Wang, S., Wang, L., Chen, W., et~al.: {LoRA}: Low-rank adaptation of large language models. International Conference on Learning Representations  \textbf{1}(2), ~3 (2022)

\bibitem{huang2021gloria}
Huang, S.C., Shen, L., Lungren, M.P., Yeung, S.: {GLoRIA}: A multimodal global-local representation learning framework for label-efficient medical image recognition. In: Proceedings of the IEEE/CVF International Conference on Computer Vision (ICCV). pp. 3942--3951 (October 2021)

\bibitem{huang2026llm2clip}
Huang, W., Wu, A., Yang, Y., Luo, X., Yang, Y., Naseem, U., Wang, C., Dai, Q., Dai, X., Chen, D., et~al.: {LLM2CLIP}: Powerful language model unlocks richer cross-modality representation. Proceedings of the AAAI Conference on Artificial Intelligence  \textbf{40}(7),  5131–5139 (Mar 2026)

\bibitem{irvin2019chexpert}
Irvin, J., Rajpurkar, P., Ko, M., Yu, Y., Ciurea-Ilcus, S., Chute, C., Marklund, H., Haghgoo, B., Ball, R., Shpanskaya, K., et~al.: {CheXpert}: A large chest radiograph dataset with uncertainty labels and expert comparison. In: Proceedings of the AAAI conference on artificial intelligence. vol.~33, pp. 590--597 (2019)

\bibitem{johnson2019mimic}
Johnson, A.E., Pollard, T.J., Berkowitz, S.J., Greenbaum, N.R., Lungren, M.P., Deng, C.y., Mark, R.G., Horng, S.: {MIMIC-CXR}, a de-identified publicly available database of chest radiographs with free-text reports. Scientific data  \textbf{6}(1), ~317 (2019)

\bibitem{lau2018VQARADdataset}
Lau, J.J., Gayen, S., Ben~Abacha, A., Demner-Fushman, D.: A dataset of clinically generated visual questions and answers about radiology images. Scientific data  \textbf{5}(1),  180251 (2018)

\bibitem{li2023llava}
Li, C., Wong, C., Zhang, S., Usuyama, N., Liu, H., Yang, J., Naumann, T., Poon, H., Gao, J.: {LLaVA-Med}: Training a large language-and-vision assistant for biomedicine in one day. In: Oh, A., Naumann, T., Globerson, A., Saenko, K., Hardt, M., Levine, S. (eds.) Advances in Neural Information Processing Systems. vol.~36, pp. 28541--28564. Curran Associates, Inc. (2023)

\bibitem{lin2023pmc}
Lin, W., Zhao, Z., Zhang, X., Wu, C., Zhang, Y., Wang, Y., Xie, W.: {{PMC-CLIP}: Contrastive Language-Image Pre-training Using Biomedical Documents}. In: Greenspan, H., Madabhushi, A., Mousavi, P., Salcudean, S., Duncan, J., Syeda-Mahmood, T., Taylor, R. (eds.) Medical Image Computing and Computer Assisted Intervention -- MICCAI 2023. pp. 525--536. Springer Nature Switzerland, Cham (2023)

\bibitem{liu2021slake}
Liu, B., Zhan, L.M., Xu, L., Ma, L., Yang, Y., Wu, X.M.: {SLAKE}: A semantically-labeled knowledge-enhanced dataset for medical visual question answering. In: 2021 IEEE 18th international symposium on biomedical imaging (ISBI). pp. 1650--1654. IEEE (2021)

\bibitem{moor2023foundation}
Moor, M., Banerjee, O., Abad, Z.S.H., Krumholz, H.M., Leskovec, J., Topol, E.J., Rajpurkar, P.: Foundation models for generalist medical artificial intelligence. Nature  \textbf{616}(7956),  259--265 (2023)

\bibitem{moor2023med}
Moor, M., Huang, Q., Wu, S., Yasunaga, M., Dalmia, Y., Leskovec, J., Zakka, C., Reis, E.P., Rajpurkar, P.: {Med-Flamingo}: a multimodal medical few-shot learner. In: Hegselmann, S., Parziale, A., Shanmugam, D., Tang, S., Asiedu, M.N., Chang, S., Hartvigsen, T., Singh, H. (eds.) Proceedings of the 3rd Machine Learning for Health Symposium. Proceedings of Machine Learning Research, vol.~225, pp. 353--367. PMLR (10 Dec 2023)

\bibitem{oquab2023dinov2}
Oquab, M., Darcet, T., Moutakanni, T., Vo, H., Szafraniec, M., Khalidov, V., Fernandez, P., Haziza, D., Massa, F., El-Nouby, A., et~al.: {DINOv2}: Learning robust visual features without supervision. arXiv preprint arXiv:2304.07193  (2023)

\bibitem{su2025large}
Su, Y., Babore, Y.B., Kahn~Jr, C.E.: A large language model to detect negated expressions in radiology reports. Journal of imaging informatics in medicine  \textbf{38}(3),  1297--1303 (2025)

\bibitem{tu2024towards}
Tu, T., Azizi, S., Driess, D., et~al.: Towards generalist biomedical {AI}. NEJM AI  \textbf{1}(3),  AIoa2300138 (2024)

\bibitem{velamala2025medspacyv}
Velamala, B., Sagheb Hossein~Pour, E., Lin, M., Fan, J.W.: {medspacyV}: a graphical user interface for the open source {medspaCy} natural language processing package. JAMIA Open  \textbf{8}(4),  ooaf094 (2025)

\bibitem{wang2022medclip}
Wang, Z., Wu, Z., Agarwal, D., Sun, J.: {MedCLIP}: Contrastive learning from unpaired medical images and text. In: Proceedings of the 2022 Conference on Empirical Methods in Natural Language Processing. pp. 3876--3887 (2022)

\bibitem{wu2024pmcllama}
Wu, C., Lin, W., Zhang, X., Zhang, Y., Xie, W., Wang, Y.: {PMC-LLaMA}: toward building open-source language models for medicine. Journal of the American Medical Informatics Association  \textbf{31}(9),  1833--1843 (2024)

\bibitem{wu2023medklip}
Wu, C., Zhang, X., Zhang, Y., Wang, Y., Xie, W.: {MedKLIP}: Medical knowledge enhanced language-image pre-training for {X}-ray diagnosis. In: Proceedings of the IEEE/CVF International Conference on Computer Vision (ICCV). pp. 21372--21383 (October 2023)

\bibitem{you2023cxr}
You, K., Gu, J., Ham, J., Park, B., Kim, J., Hong, E.K., Baek, W., Roh, B.: {CXR-CLIP}: Toward large scale chest {X}-ray language-image pre-training. In: International Conference on Medical Image Computing and Computer-Assisted Intervention. pp. 101--111. Springer (2023)

\bibitem{BiomedCLIP}
Zhang, S., Xu, Y., Usuyama, N., Xu, H., Bagga, J., Tinn, R., et~al.: A multimodal biomedical foundation model trained from fifteen million image–text pairs. NEJM AI  \textbf{2}(1),  AIoa2400640 (2025)

\bibitem{zhang2022contrastive}
Zhang, Y., Jiang, H., Miura, Y., Manning, C.D., Langlotz, C.P.: Contrastive learning of medical visual representations from paired images and text. In: Machine learning for healthcare conference. pp. 2--25. PMLR (2022)

\bibitem{zhao2025clip}
Zhao, Z., Liu, Y., Wu, H., Wang, M., Li, Y., Wang, S., Teng, L., Liu, D., Cui, Z., Wang, Q., et~al.: {CLIP} in medical imaging: A survey. Medical Image Analysis  \textbf{102},  103551 (2025)

\bibitem{MMedPO_ICML25}
Zhu, K., Xia, P., Li, Y., Zhu, H., Wang, S., Yao, H.: {MMedPO}: aligning medical vision-language models with clinical-aware multimodal preference optimization. In: Proceedings of the 42nd International Conference on Machine Learning (ICML). ICML'25, JMLR.org (2025)

\end{thebibliography}
\end{document}